# A NOVEL APPROACH FOR GLAUCOMA CLASSIFICATION BY WAVELET NEURAL NETWORKS USING GRAPH-BASED, STATISITCAL FEATURES OF QUALITATIVELY IMPROVED IMAGES

N. Krishna Santosh, Dr. Soubhagya Sankar Barpanda
School of Computer Science Engineering, VIT - AP University, India 522237

## Abstract

In this paper, we have proposed a new glaucoma classification approach that employs a wavelet neural network (WNN) on optimally enhanced retinal images' features. To avoid tedious and error-prone manual analysis of retinal images by ophthalmologists, computer-aided diagnosis (CAD) substantially aids in robust diagnosis. Our proposal's objective is to introduce a CAD system with a fresh approach. Retinal image quality improvement is attempted in two phases. The retinal image preprocessing phase improves the brightness and contrast of the image through quantile-based histogram modification. It is followed by the image enhancement phase, which involves multi-scale morphological operations using image-specific dynamic structuring elements for the retinal structure enrichment. Graph-based retinal image features in terms of Local Graph Structures (LGS) and Graph Shortest Path (GSP) statistics are extracted from various directions along with the statistical features from the enhanced retinal dataset. WNN is employed to classify glaucoma retinal images with a suitable wavelet activation function. The performance of the WNN classifier is compared with multilayer perceptron neural networks with various datasets. The results show our approach's superior performance to the existing approaches.

***Key words:*** *Retinal image improvement, dynamic structuring element, local graph structures, graph shortest path, wavelet neural networks.*

## 1. Introduction

In human life, vision plays a critical part in coordinating daily activities. There are harmful diseases that can affect vision more than the adverse effects caused by accidents. The primary cause [1] of vision disability is the development of glaucoma. There are a variety of approaches to detecting glaucoma in the early stages [2], most of which are based on medical examination through the fundus (retinal) image structures [3]. Manually inspecting retinal images, on the other hand, is a time-consuming and error-prone task. In this case, CAD [4] has proven to be a valuable tool for ophthalmologists in interpreting the eye fundus image for a rapid and accurate glaucoma diagnosis. The major phases of CAD are image preprocessing, feature extraction, and classification. In this paper, we have proposed a novel CAD system whose significant contributions are utilizing quantile-based preprocessing for mapping image histograms, and complex wavelet-based image enhancement using dynamic structuring elements. It is followed by a powerful feature extraction phase that considers both statistical and graph-based features, which are employed for retinal image classification using WNN and multilayer perceptron neural networks (MLP).

The rest of the paper is organized into different portions: Section 2 gives the existing significant contributions and motivation, Section 3 illustrates the proposed method of retinal image enhancement, feature extraction techniques, Section 4 describes about the implementation of classification, Section 5 analyzes the classification results for performance appraisal, and Section 6 concludes the paper.

## 2. Literature survey and motivation

In the proposed method, our contribution exists in retinal image enhancement, novel feature extraction, and classification. So far, numerous practices are proposed in all these phases of the CAD system. Out of all the contributions, some significant contributions are discussed below.

It is essential to improve the visual quality of retinal image structures for effective feature extraction. Gupta et al. [5], Kim et al. [6], and Sim et al. [7] employ adaptive gamma correction (AGC) and histogram equalization (HE) approaches to enhance the luminosity of the image without the occurrence of a gamut problem. Histogram modification based on recursive procedures, optimization is given by Wang et al. [8], Arici et al. [9], and Chen et al. [10], but these approaches are not optimal due to their recursive nature. A weighted distribution is applied to image regions through AGC by Cheng et al. [11]. A discrete wavelet transformation (DWT) is applied by Mallat et al. [12]. DWT limitations are addressed by Selesnick et al. [13] via proposing a complex wavelet transformation (CWT) based image enhancement. The contourlet transform (CT) of the retinal image was proposed by Peng et al. [14], using non-linear functions. Morphological image enhancement was introduced by Andrew et al. [15] to prevent the occurrence of spurious details and the identification of sharp edges. Liao et al. [16] applied a top-hat transformation (TH) with multi-scale to improve the retinal image based on brighter and dim regions.

In our proposed method, statistical and graph-based image features are considered. The co-occurrence matrix of image gray level (GLCM) is used by Giraddi et al. [17] to extract the features of second-order. Wavelet features are extracted from the image by Sumeet et al. [18] to classify images using Naïve Bayes (NB) [32], Support Vector Machine (SVM) [29], and Random Forest (RM) [31] approaches. Ranked cross-entropy image features are utilized by Shubhangi et al. [19]. Statistical image features are employed by Anindita et al. [20] for higher accuracy. Local binary pattern (LBP) based features are extracted by Jefferson et al. [21] for glaucoma classification. Anushikha et al. [22] extracted DWT features from the vessel free optic disc area to get higher accuracy in ANN [28] classification. A hybrid feature set is formed with structural and non-structural features by Anum et al. [23]. Images' higher-order spectra (HOS) features are optimized by linear discriminant analysis (LDA) prior to applying SVM, and NB classifiers by Kevin et al. [24] and Acharya et al. [25]. Image texture features can also be extracted from the equivalent graphs obtained from fundus images, which comes under graph-based features. Abusham et al. [26] formed a binary relationship pattern for texture features using the graph formed by image mapping. Mesquita et al. [27] extracted the shortest paths from the image equivalent graph to generate texture features for image categorization. The nearest neighbor classifier (k-NN) is employed by Anindita et al. [30] on statistical retinal image features for higher accuracy.

It has been observed from the existing literature that prior to feature extraction, image quality improvement is carried out either by focusing on image brightness and contrast, or on image structures, but not both. Regarding feature extraction, there is no specific approach to capturing the image texture pattern. This motivated us to propose a fresh approach for the image quality enhancement, feature extraction, and classification phases of the CAD-based glaucoma detection. Concerning retinal image quality, both color adjustment and retinal structural improvement are necessary. Thus, we have incorporated two sub-phases into retinal image quality improvement to achieve intensity adjustment and structural improvement. The CAD approach to image classification is heavily based on the texture patterns of image. This led us to identify a new approach called graph-based image feature extraction, along with statistical features for glaucoma classification. We have also introduced wavelet neural networks (WNN) for glaucoma identification. As per the available literature, our proposed approach is the first one to use WNN for glaucoma classification.

## 3. Proposed CAD approach for glaucoma diagnosis

Our CAD approach is carried out on the ORGIA dataset [33], which is publicly available and results are compared with the other public datasets (ACRIMA and DRISHTI). The overall flow of the proposed method of CAD glaucoma diagnosis is shown in Fig. 1. The proposed process begins with retinal image preprocessing followed by enhancement for the improved visual appearance of images, which is the best seed for significant feature extraction. Each image feature vector is fed to ML classifiers to diagnose the image type (normal or abnormal). Classifier findings are analyzed to get a meaningful conclusion. This section explains the proposed image preprocessing, enhancement, and feature extraction w.r.t corresponding algorithm step numbers ($An:Sn$).

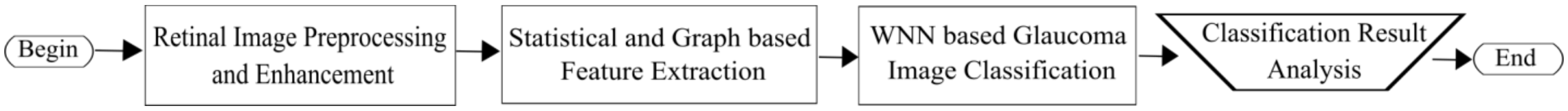

**Fig. 1: High level process flow of the proposed method**

### 3.1. Retinal image pre-processing

The proposed preprocessing qualitatively improves the retinal image by systematically increasing brightness and contrast, as given in algorithm 1. It begins by formalizing the gray image's histogram modifications into an optimization problem to make them closer to the ideal histogram ($A1:S3$). Using the resultant mapped histogram (*Mh*), a normalized cumulative distribution function ($F_{CDF}$) is generated ($A1:S6$). Instead of conventional $F_{CDF}$ usage [5], we have employed a quadratic rank transmutation map (QRTM) [34] transformation of CDF ($CDF_T$) to provide a skew-kurtotic normal distribution ($A1:S8$). Modification to the AGC [35] is carried out to form a brighter image ($I_{AGC}$) ($A1:S9$). During this process, the color information of the image may slightly fluctuate, which can be restored by the input retinal image ($I_{ORG}$) color information ($A1:S10$). The resultant $IMG_{pre}$ is a luminous and brightness enhanced image now subjected to contrast improvement using a quantile-based approach. In terms of image intensities, quantiles are intensity values generated using the PDF of the image histogram ($H_{pre}$) over [$I_{min}$, $I_{max}$]. In our approach, we generated $t$-quantiles for $IMG_{pre}$ by histogram splitting, as $H_{pre1} = [i_0, i_1], H_{pre2} = [i_1, i_2], \ldots, H_{pre-t} = [i_{t-1}, i_t]$ and the corresponding normalized $CDF_{pre\text{-}k}$ for each sub-histogram is generated. The final pre-processed image ($IMG_{PRE}$) is obtained using sub-histogram mapping ($A1:S12$). The resultant preprocessed retinal images are enriched with sufficient brightness and contrast, and the sample results are shown in Fig. 3. These preprocessed retinal images are fed into the image enhancement process.

**Algorithm 1** Retinal image pre-processing

**Input:** Gray scale retinal images: $GI$

**Output:** Brightness and contrast improved retinal image: $IMG_{PRE}$

1: **procedure** PRE-PROCESS($GI$)
2:     **while** $GI \in DATASET$ **do**
3:         $Mh \leftarrow \alpha_{opt} HIST_{GI} + (1 - \alpha_{opt}) HIST_U$
4:
5:         **for** <$I \in GM$ from 0 to $I_{\max}$> **do**
6:             $F_{CDF}(I) \leftarrow \sum_{l=0}^{I-1} \left( \frac{Mh}{\sum_{l=0}^{I-1} Mh(l)} \right)$
7:         **end for**
8:         $CDF_T \leftarrow (1+\delta) F_{CDF} - \delta (F_{CDF})^2$
9:         $Generate : I_{AGC} \leftarrow (I_{max} - 1) \left( \frac{I}{I_{max}} \right)^{1 - CDF_T(I)}$
10:         $Color\,restoration : IMG_{pre} \leftarrow \vartheta (I_{AGC}) + (1 - \vartheta)(I_{ORG})$.
11:         **for** <$(k\, from\, 1\, to\, t)$ $and$ $(i\, from\, 0\, to\, \max - 1)$> **do**
12:             $IMG_{PRE} = \bigcup i_{k-1} + (i_k - i_{k-1}) CDF_{pre-k} [I_i]$
13:         **end for**
14:     **end while**
15: **end procedure**

### 3.2. Retinal image enhancement

The proposed enhancement technique aims to increase the quality of retinal image structures as described in algorithm 2. Dual tree CWT (DTCWT) high and low-pass sub-bands [36, 37] are obtained from $IMG_{PRE}$. After this, the low and high pass sub-bands are processed independently as described in the following sub-sections.

#### 3.2.1. Low-pass coefficient enhancement using dynamic structuring element based morphological operation

Our method uses an improved top-hat transformation (TH) using white ($TH_W$) and bottom top-hat ($TH_B$) operations for the low-pass image coefficient mapping. Usage of the same structuring element (SE) results in poor utilization of pixel neighboring information. To address this, our proposed approach performs TH on multiple scales with varied $SE$ for processing brighter and darker image areas. To enhance the structural components of the image, pixel gray value transformation is one of the finest solutions [38]. Our approach defines *SE* dynamically ($D_{SE}$) based on the s-curve optimization principle (illustrated in Fig. 2), that transforms larger gray pixel values into small higher ranges and smaller gray pixel values into short-ranges.

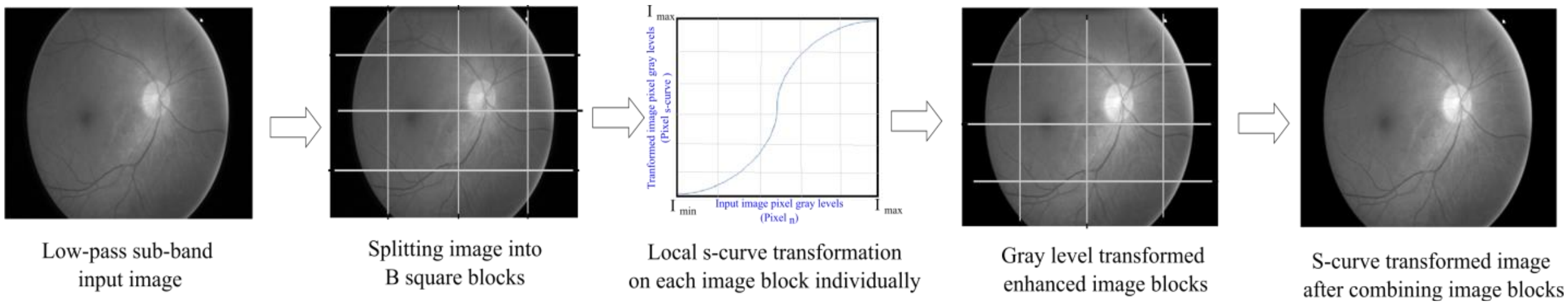

**Fig. 2 Local s-curve image transformation procedure**

We have constructed the $D_{SE} = \{SE_1, SE_2, \ldots SE_i, \ldots SE_l\}$ using the initial $SE_0$ ($A2:S23$). The $i^{th}$ *SE* is dynamically constructed by operating on $SE_0$ for t-times, its value being determined by using local s-transformation. In the initial iteration, $SE_i$ is obtained with $t = 1$. It is followed by *LS* enhancement ($LS_{en}$) ($A2:S24$). Local s-transformation is operated on $LS_{en}$ and the edge content [39] of the resulted s-curve transformed image ($ED_{C_LS}$) is measured ($A2:S25$ *to* $S28$). The distance to the original image edge content value ($ED_C$) is the deciding factor of *t* value and thus $SE_i$ ($A2:S29$ *to* $S32$). At the end of iterations, the *t* value is determined for dynamic structuring element $D_{se}$. Using $D_{se_i}$ in $TH_{W_i}$ and $TH_{B_i}$,the low-pass sub-band image is enhanced ($LS_{final}$) by upgrading brighter ($DTH_W$) and darker ($DTH_B$) regions at equal rates using control limit (*K*) based on the edge detection principle ($A2:S6$ *to* $S8$).

### 3.2.2. High-pass coefficient enhancement using contourlet transform

The process starts by applying contourlet transform (*CT*) on retinal images' high-pass sub-band (HS) DTCWT coefficients to eliminate noise components [40]. The noise-free HS image ($HS_{Noise\text{-}free}$) is generated using noise variance ($Noise_{SB}$) and variance of sub-bands ($Variance_{SB}$) of size $p \, X \, q$ w.r.t. $A \, X \, A$ sized neighbor area (*A2:S9 to S13*).

### 3.2.3. Formation of enhanced retinal image

Inverse DTCWT is applied to combine the processed DTCWT *LS* and *HS* of the retinal image to produce a fully improved retinal image ($IMG_{enh}$) (*A2:S14*). The process corrects the over brightness and contrast in the enhanced retinal images, and the significant retinal structures are visually highlighted, as shown in Fig. 3. The final enhanced images are utilized for feature extraction, which is described in the next sub-section.

**Algorithm 2** Retinal image enhancement

**Input:** `Gray scale retinal images:`$GI$

**Output:** `Brightness and contrast improved retinal image:`$IMG_{PRE}$

**procedure** PRE-PROCESS($GI$)
  **while** $IMG \leftarrow IMG_{PRE}, IMG \notin Empty$ **do**
    $\{LS_{set}, HS_{set}\} \leftarrow DTCWT\,(IMG)$
    **for** <$LS \in \{LS_{set}\}$> **do**
      $D_{se} \leftarrow$ *call* $D_{se}$ *construction*
      $TH_{W_i}(m,n) = LS(m,n) - LS \circ D_{se_i}(m,n)$ ; $TH_{B_i}(m,n) = LS \bullet D_{se_i}(m,n) - LS(m,n)$
      $DTH_{w_i}(m,n) = TH_{W_{i+1}}(m,n) - TH_{W_i}(m,n)$ ; $DTH_{B_i}(m,n) = TH_{B_{i+1}}(m,n) - TH_{B_i}(m,n)$
      $LS_{final} = LS + 1 - \kappa_i$
    **end for**
    **for** <$HS \in \{HS_{set}\}, HS \notin Empty$> **do**
      $SB \leftarrow CT\,(HS)$
      $HS_{Noise-free} \leftarrow \frac{1}{p\times q}\sum_{r,s=1}^{r=p,s=q}\left(\frac{1}{p\times q}\sum_{r,s=1}^{r=p,s=q}\left(\frac{1}{A^2}\sum_{a_1,a_2=1}^{A}\left|SB\,(a_1,a_2)\right|^2\right)\right) - \frac{Median(|C_{SB}|)}{\beta}$
    **end for**
    $IMG_{enh} \leftarrow DTCWT_{Inverse}\,(LS_{final}, HS_{Noise-free})$
  **end while**
**end procedure**
**procedure** $D_{se}$ CONSTRUCTION($IMG_{PRE}$)
  **for** <$IMG \in IMG_{PRE}$> **do**
    $\{LS\} \leftarrow DTCWT\,(IMG)$
    $t \leftarrow 1$
    **for** <$LS \in \{LS\}$> **do**
      **while** $(TRUE)$ **do**
        $SE_i \leftarrow SE_0 \oplus SE_0 \oplus ... \oplus SE_0 \; for\, t - times$
        $LS_{en} \leftarrow LS + [LS(m,n) - LS \circ SE_i(m,n)] - [LS \bullet SE_i(m,n) - LS(m,n)]$
        **for** <$pixel \in LS_{en}$> **do**
          $G_{LS_{en}} \leftarrow C + \frac{R}{1+\exp\left(\frac{Pixel-\delta_1}{\delta_2}\right)}$
        **end for**
        $ED_{C_LS} \leftarrow \frac{1}{M*N}\sum_{m=1}^{M}\sum_{n=1}^{N}\left|G_{LS_{en}}\,(m,n)\right|$
        **if** $(|ED_C - ED_{C_LS}| \leq Diff_{\max})$ **then**
          $t \leftarrow t + 1$
        **else**
          $return\, D_{se} \leftarrow SE_i$
        **end if**
      **end while**
    **end for**
  **end for**
**end procedure**

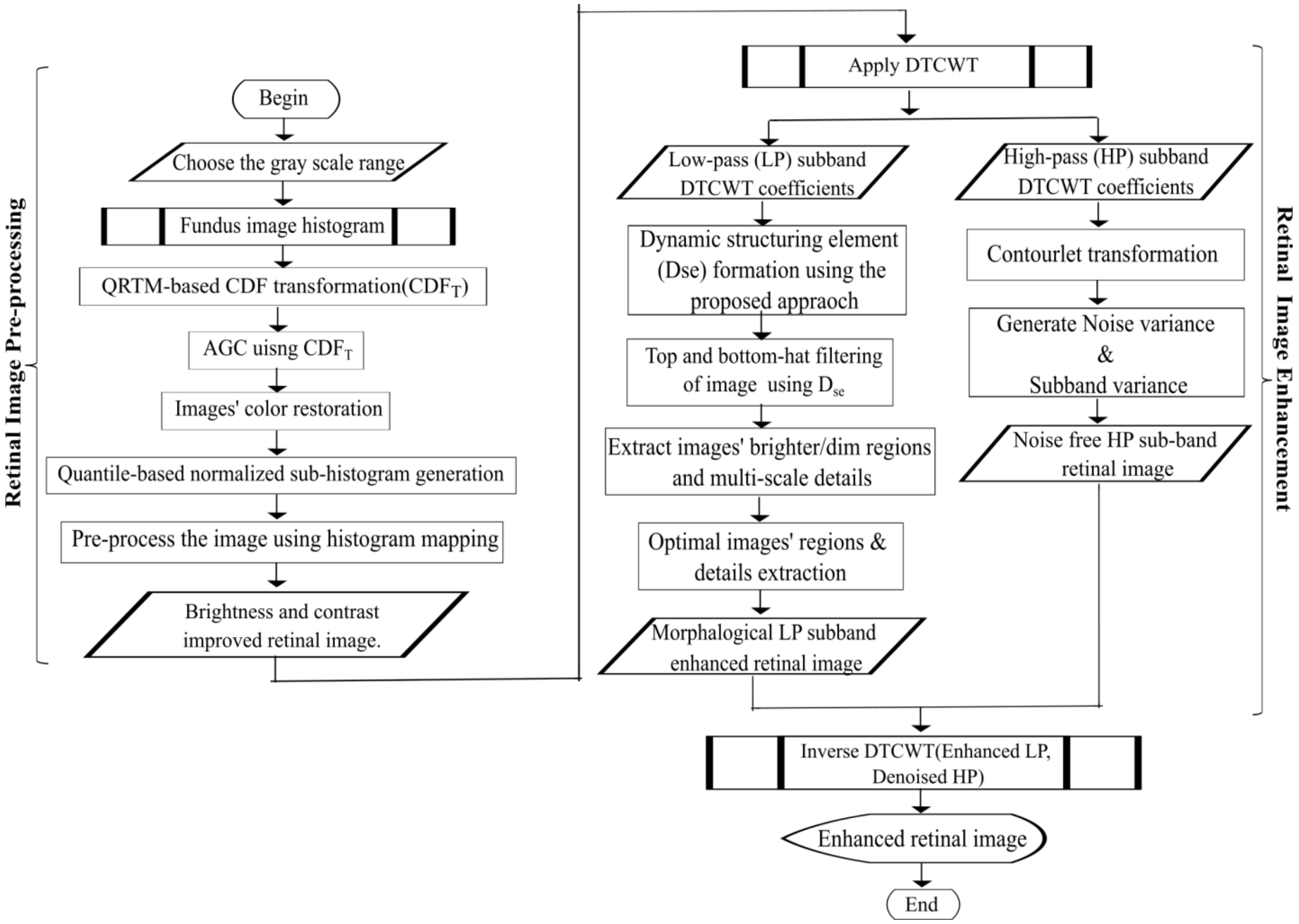

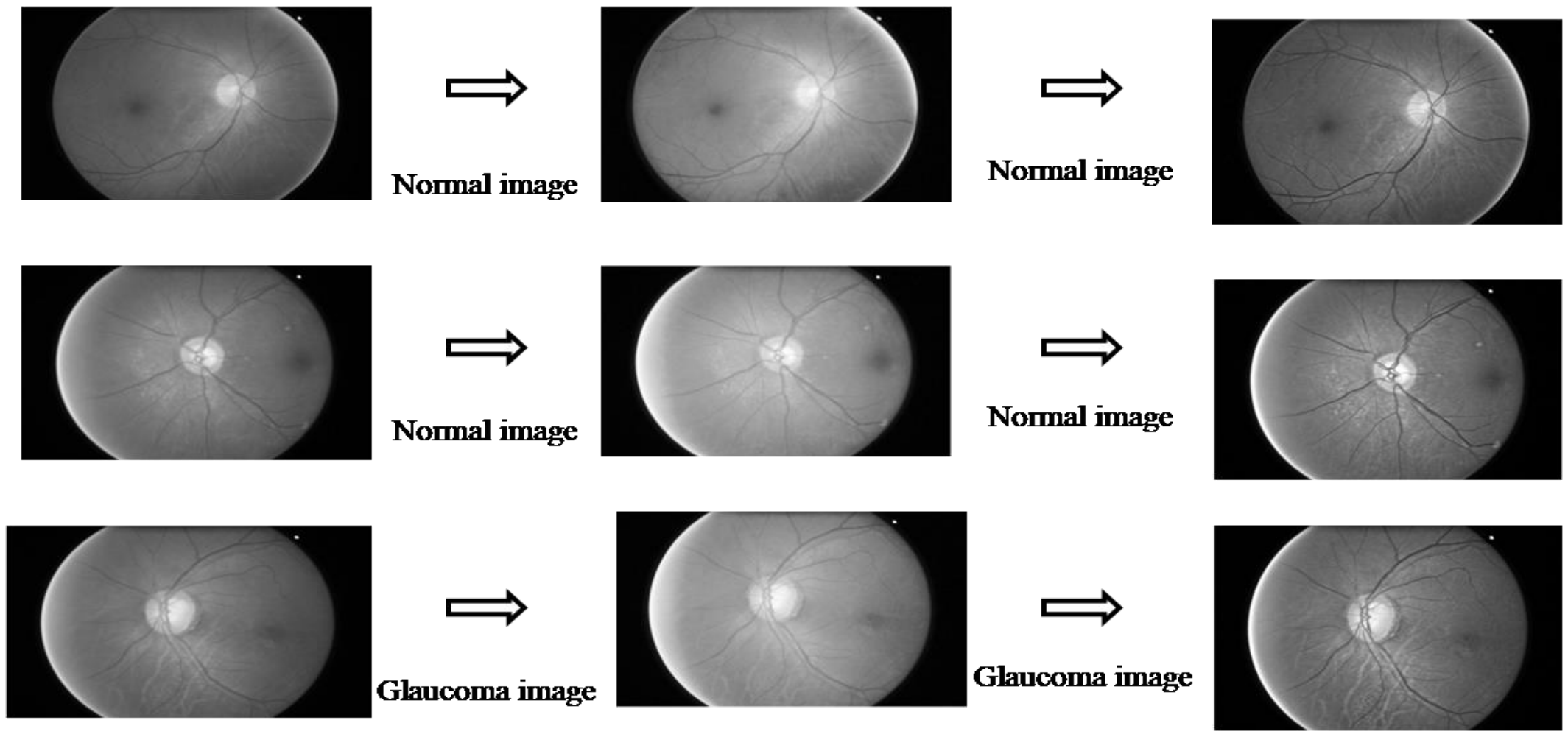

**(a) Original image** **(b) Preprocessed image** **(c) Enhanced image**

**Fig. 3: Sample results of proposed retinal image enhancement**

### 3.3. Retinal image feature extraction

The core part of the classification process is the extraction of prominent features, which is both tricky and sensible. The overall feature extraction process is described in algorithm 3. We have considered statistical and graph-based retinal image features from enhanced images to form a feature vector.

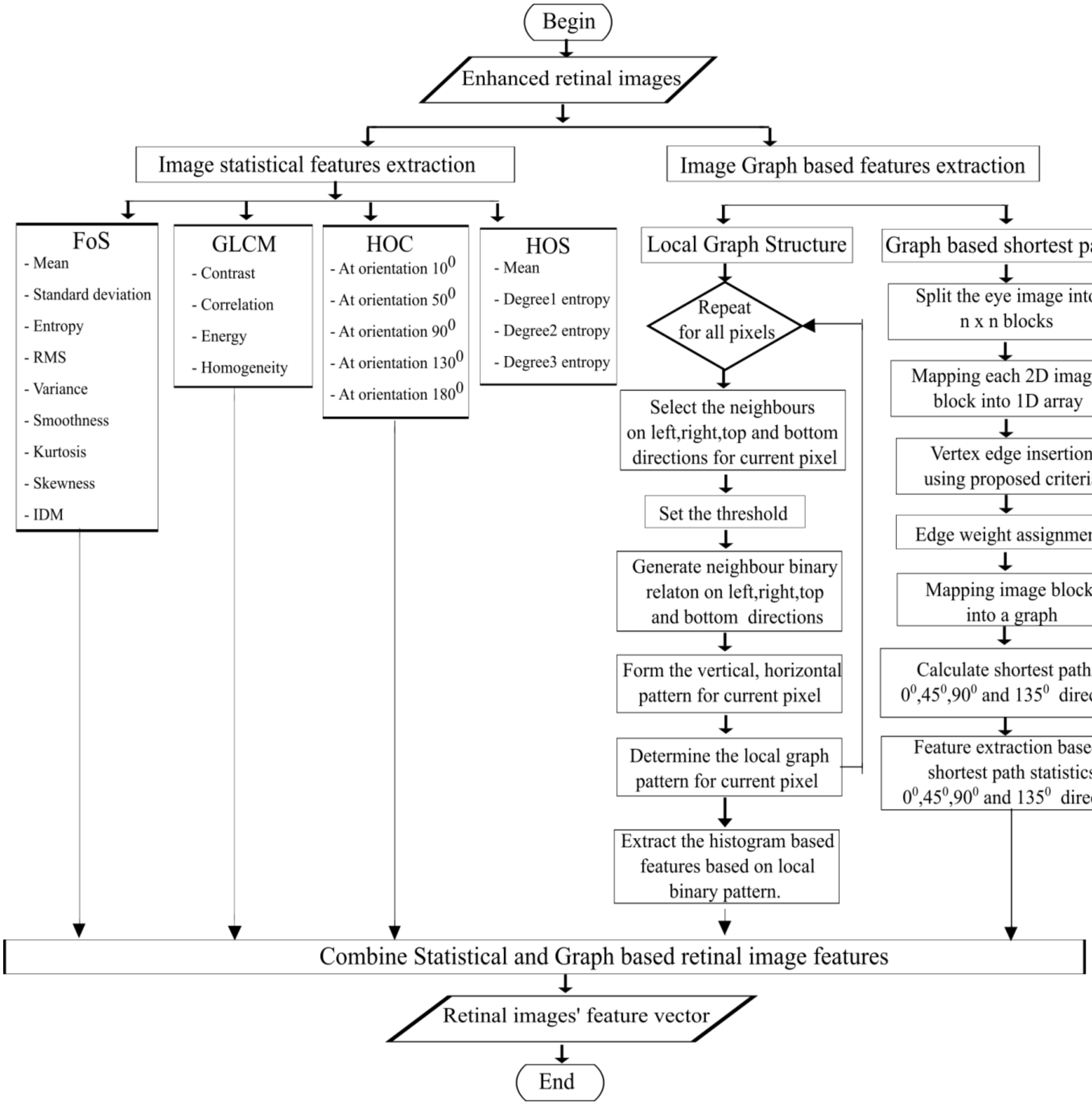

#### 3.3.1. Retinal images' statistical feature extraction

On the enhanced retinal image, we applied DWT using a two-channel filter bank [41, 42]. In this work, we have considered the biorthogonal 6.8 (Bior 6.8) wavelet filter banks. The resultant coefficients (*Cf*) are used to extract first and second order statistical (FoS, SoS) [43] features. Retinal image characteristics that fall into the non-stationary and non-linearity categories are captured with HOC features [44], which capture the higher-order correlations of input signals that are generated from moments of higher order. Since retinal images come under the non-stationary signal category, in our approach, third-order cumulants in various directions of $10^0$, $50^0$,$90^0$, $130^0$, and $180^0$ are considered. Signal amplitude and phase information are used for HOS feature extraction such as mean and entropies of different degrees [45]. The characteristics of non-Gaussian, non-stationary in image are captured by HOS features.

### 3.3.2. Retinal images' graph-based feature extraction

Graph representations of images provide a structure for pictorial data encoding [46]. Each pixel of the image is interpreted as a node in the graph, and connections to the neighboring pixels are represented by edges. The formed graph reflects the image's hidden patterns through its structures, which are highly significant in image description and identification [47]. In our approach, local graph structures and shortest paths are considered for graph-based image feature extraction.

**A) Local graph structures (LGS) features**

LGS feature extraction operates on a graph theory that considers pixels as vertices ($Vertex_G$) and their neighborhood relations. It applies the threshold operation on neighboring pixels based on current pixels in different directions. To avoid the imbalanced neighbor covering limitation with asymmetric LGS, in our approach, a symmetrically shaped neighborhood is considered for every pixel point in the graph. The enhanced retinal image pixel gray intensities are considered as points on the graph. In the earlier work, only the left and right regions of the pixel were considered for forming a symmetric region [48] that represents the pixel's neighbor relationship. In our approach, the top and bottom neighbor regions are considered along with the left and right regions to form symmetric regions for every pixel considered as the current pixel. Symmetric regions are considered for the immediate left, right, top, and bottom neighbors of the current pixel in a predefined order indicated in Fig. 4.a.

The LGS operator compares pair-wise pixel gray values starting from the center pixel in the given directions. While moving a pair of points (direction shown by arrow), the connecting edge is assigned a label of *zero* in the case of a higher to lower intensity value, otherwise assigned a label of *one*. This process continued for every pair of vertices of the central pixel neighborhood symmetric region (SNR). Following this, two binary patterns are generated. The first pattern ($Pattern_{LR}$) is generated by considering all the resultant binary threshold bits from the left and right regions of the central pixel. The second one ($Pattern_{TB}$) is generated by considering the top and bottom regions. Once two patterns with a $P$ number of bits are obtained, the magnitude of the neighbor relation pattern (NRP) results from the threshold operation ($Thr$) being generated using Eq. 1.

$$NRP_{LR} = \sum_{p=0}^{P-1} Thr(C_V, \mathrm{HN}_V) 2^p \quad ; \quad NRP_{TB} = \sum_{p=0}^{P-1} Thr\left(C_V, VN_V\right) 2^p \tag{1}$$

A threshold operation is applied on every pair of central vertex ($C_V$) and horizontal or vertical neighbor vertices ($HN_V$ or $VN_V$) that results in two magnitude values for the left to right ($NRP_{LR}$) and top to bottom directions ($NRP_{TB}$) of the center vertex ($C_V$). The final $NRP$ ($NRP_{Final}$) for every pixel point in the graph is generated using Eq. 2.

$$NRP_{Final} = \sqrt{\left(NRP_{LR}\right)^2 + \left(NRP_{TB}\right)^2} \tag{2}$$

The resultant value of $NRP_{Final}$ is used to replace the value of the current pixel point as shown in Fig. 4.b. A histogram ($H_{LGS}$) is generated for the resultant LGS retinal image of size $N_r \times N_c$. Probability density is calculated approximately using Eq. 3.

$$\text{Prob}_{LGS} = {}^{H_{LGS}}\!/_{N_r * N_c} \tag{3}$$

Using $Prob_{LGS}$, the histogram features for every intensity level ($L$) are generated using Eq. 4 to form the LGS image feature vector.

$$Mean_{LGS} = \sum_{L=0}^{L-1} \text{Prob}_{LGS_L}; Variance_{LGS} = \sum_{L=0}^{L-1} \left(\text{Prob}_{LGS_L}\right)^2; \ Skewness_{LGS} = \left(\sqrt{Variance_{LGS}}\right)^{-3} * \sum \text{Prob}_{LGS};$$

$$Kurtosis_{LGS} = \left(\sqrt{Variance_{LGS}}\right)^{-4} * \sum \text{Prob}_{LGS} \ ; \ Energy_{LGS} = \sum_{L=0}^{L-1} \left(\text{Prob}_{LGS_L}\right)^2;$$

$$Entropy_{LGS} = -\sum_{L=0}^{L-1} \left(\text{Prob}_{LGS_L}\right) * \log\left(\text{Prob}_{LGS_L}\right) \tag{4}$$

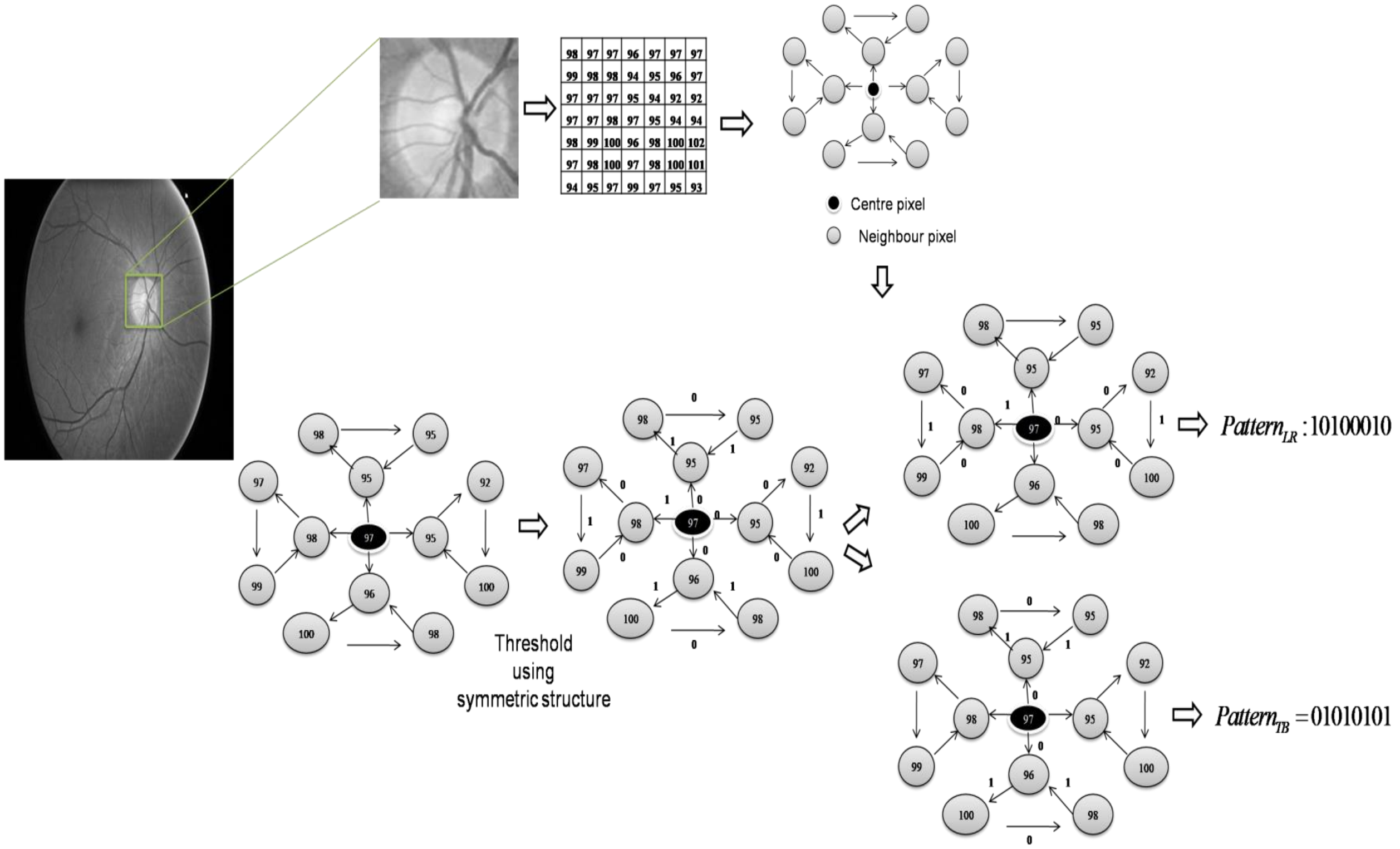

**(a) Neighbor relation pattern using LGS operation**

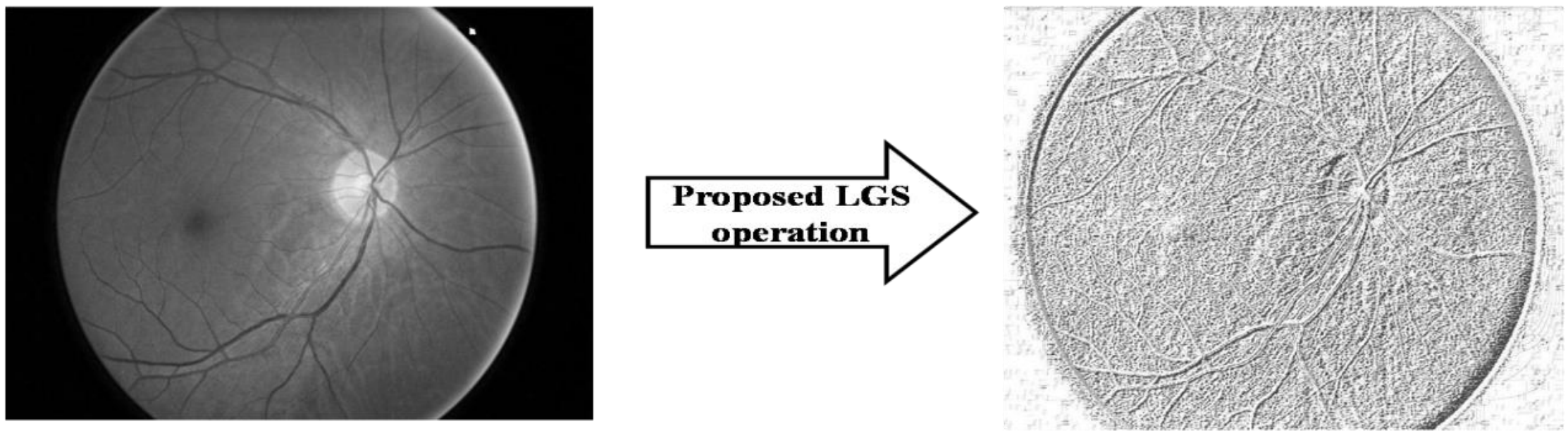

**(i) Enhanced retinal image** **(ii) Resulted LGS retinal image**

**(b) Sample result of proposed LGS**

**Fig. 4 Local graph structures feature extraction**

### B) Graph based shortest path (GSP) features

Our proposed GSP approach is a powerful tool that explores retinal image textures using shortest path statistics between different pixel points in various orientations. A graph *G (V, E)* is a novel structure for representing data using vertices $v \in V$ and the relationships between data points using edges $e \in E$. The weights of each edge represent the significance of the vertex pair relationship. The approach begins with mapping the retinal image into an undirected weighted graph. Each pixel gray value is treated as a vertex of the graph. Initially, the 2D gray retinal image is converted into 1D data that represents the vertices *V* in the resultant retinal image graph. Following this, an undirected edge (*e*) is inserted between pairs of vertices $v_1, v_2 \in V$ based on their Chebyshev distance using Eq. 5 to form an edge set *E*.

$$E = \left\{ e = (v_1, v_2) \in A_M \left| \max\left( |x_1 - x_2|, |y_1 - y_2| \right) = T_e \right| \right\} \tag{5}$$

The edge insertion is represented by values between vertex pairs using an adjacent matrix ($A_M$). In our approach, two vertices ($x_1, y_1$) and ($x_2, y_2$) are connected using an undirected edge when their Chebyshev distance is $T_e$. The graph-based shortest path texture descriptor is primarily based on the shortest path between specified vertex points, so edge weights play a significant role in this process. In this approach, edges are assigned weights using Eq. 6.

$$E_w = abs\left( I_{v_1}(x_1, y_1) - I_{v_2}(x_2, y_2) \right) + \frac{I_{v_1}(x_1, y_1) + I_{v_2}(x_2, y_2)}{2} \tag{6}$$

where $I_{v_1}(x_1, y_1)$ and $I_{v_2}(x_2, y_2)$ represent gray image intensities at specified vertex locations. The edge weight ($E_w$) is a combination of the cost of moving from $v_1$ to $v_2$ and the transition altitude in the shortest path finding. To capture the local and global texture patterns of the retinal image, we split the image into multiple equal-sized ($N_B$) blocks. This makes it possible to analyze and capture the retinal image texture more locally by forming the shortest path-based texture descriptors. The whole retinal image is split into image blocks ($I_B$) and corresponding graphs are generated using the above-described process. The shortest paths are generated between a pair of source ($S_v$) and destination ($D_v$) vertices in the graph using Dijkstra's algorithm.

To capture the in-depth retinal image texture characteristics, in our approach, four different directions are considered for each block of the image to find the shortest path. In each image block, four pairs of vertices are selected to represent the horizontal direction ($0^0$), the vertical direction ($90^0$), and both diagonal directions ($45^0$ and $135^0$) to find the shortest paths as shown in Fig. 5a. Using pixel intensities along the shortest paths, the image features are extracted as illustrated in Fig. 5b. Each image block produces four paths, $P_0$, $P_{90}$, $P_{45,}$ and $P_{135}$ for four directions, including all the intermediate vertices for path construction. Then each shortest path vertices are mapped into corresponding gray level intensities. Thus, each path $P_i$ in any one of four directions is now a collection of pixel intensities that appear in its construction. The corresponding mean ($P_{i\text{-}mean}$) is calculated for each path $P_{i,}$ $i \in \{0^0, 45^0, 90^0, 135^0\}$. As the original image is divided into $N_B$ blocks, every block generates a corresponding $Pi_{\text{-}mean.}$ Using this, four *1D* vectors ($Mean_{0^0}, Mean_{45^0}, Mean_{90^0}, Mean_{135^0}$) of size $N_B$ are formed by grouping the $Pi_{\text{-}mean}$ corresponding to each direction of every block of a retinal image. Using each $Mean_{i^0}$, the statistical measures kurtosis, skewness, standard deviation, and quantiles(Q25, Q50, Q75, Q135 ) are generated. By comparing quantiles of one data set with quantiles of another data set, they can be grouped into similar or dissimilar categories [49].

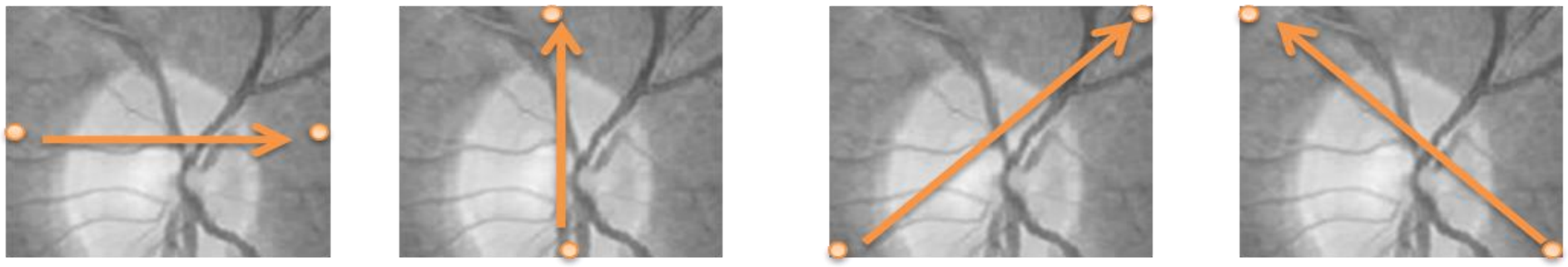

**(i) Horizontal path($0^0$)** **(ii) Vertical path ($90^0$)** **(iii) Diagonal path($45^0$)** **(iv) Diagonal path ($135^0$)**

**(a) Four directional paths for retinal image**

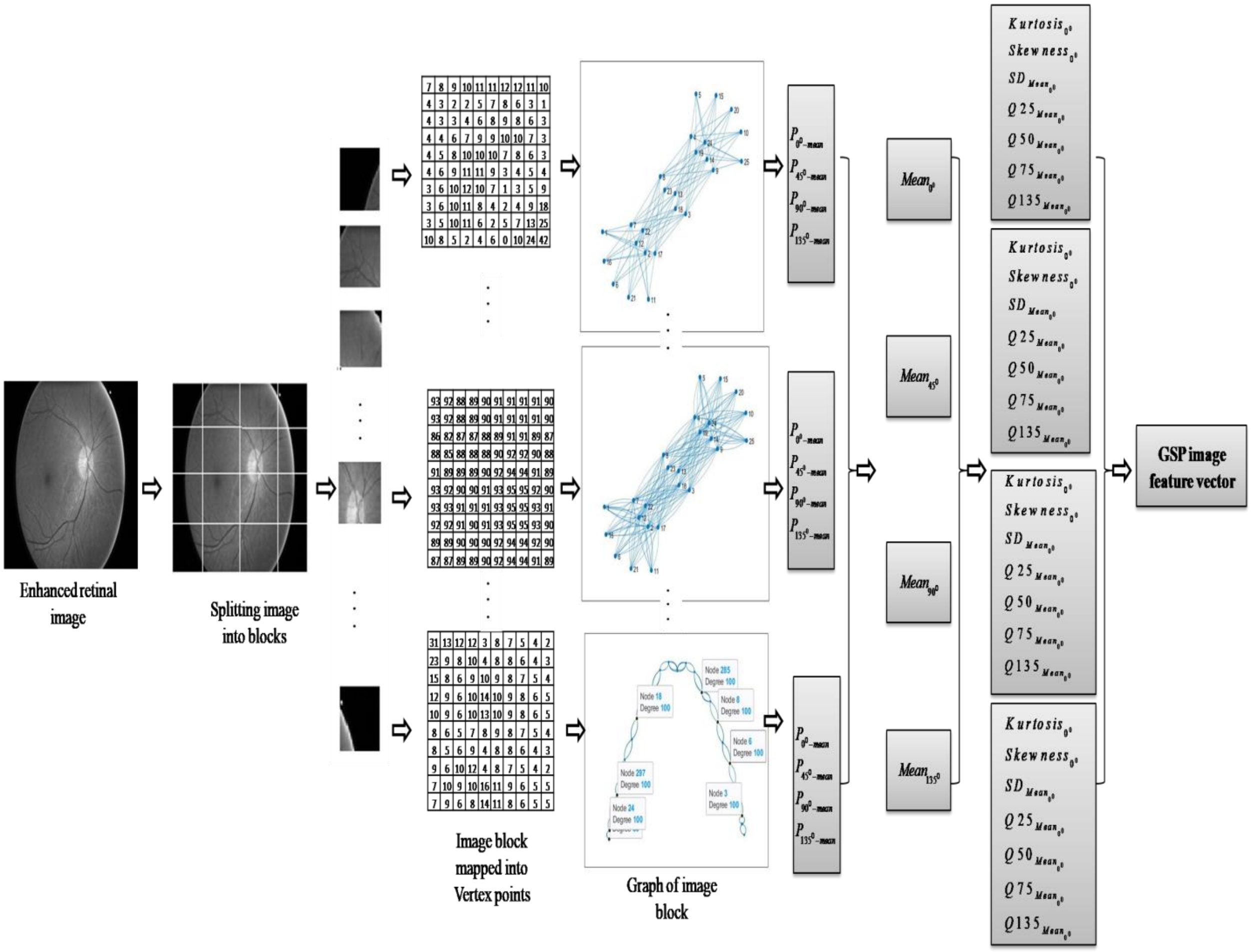

**(b) GSP feature vector generation**

**Fig. 5 Graph based shortest path retinal image feature extraction**

**Algorithm 3** Retinal image feature extraction

**Input:** `Enhanced retinal images`

**Output:** `Retinal images statistical and graph-based feature vectors`

**procedure** FEATURE EXTRACTION($IMG_{enh}$)
  **for** <$IMG \in IMG_{enh}$> **do** ▷ Image statistical features
    $Cf \leftarrow DWT_{Level-2}\,(IMG)$
    **for** <$each\,Cf$> **do**
      $FoS_{IMG} \leftarrow \{Mean\,(Cf),\ SD\,(Cf),\ Ent\,(Cf),\ RMS\,(Cf),\ Var\,(Cf),\ Smh\,(Cf),\ Kur\,(Cf$
      $SoS_{IMG} \leftarrow \{GLCM\,(Cf)\}$
      $HoC_{IMG} \leftarrow \{HOC_{10^0}\,(Cf)\,, HOC_{50^0}\,(Cf)\,, HOC_{90^0}\,(Cf)\,, HOC_{130^0}\,(Cf)\,, HOC_{180^0}\,(Cf)\}$
      $HoS_{IMG} \leftarrow \{Entropy_{HoS}\,(Cf)\,, Mean_{HoS}\,(Cf)\,, Ent_{dg1}\,(Cf)\,, Ent_{dg2}\,(Cf)\,, Ent_{dg3}\,(Cf)\}$
    **end for** ▷ Image graph-based features
    $\{Vertex_G\} \leftarrow Graph_Mapping(IMG)$
    **for** <$each\ \ V \in \{Vertex_G\}$> **do**
      $\{HN_V, VN_V\} \leftarrow SNR(C_V) \Leftarrow C_V \leftarrow V$
      **for** <$each\,p \leftarrow 0\,to\,P-1$> **do**
        $NRP_{LR} \leftarrow \sum_{p=0}^{P-1} Thr(C_V, HN_V)2^p$
        $NRP_{TB} \leftarrow \sum_{p=0}^{P-1} Thr\,(C_V, VN_V)\,2^p$
      **end for**
      $NRP_{Final} \leftarrow \sqrt{(NRP_{LR})^2 + (NRP_{TB})^2}$
      $MG_{LGS} \leftarrow Graph_Remapping\,(IMG(V) \Leftarrow NRP_{Final})$
      $LGS_{IMG} \leftarrow \{Mean\,(IMG_{LGS})\,,\ Variance\,(IMG_{LGS})\,Skewness\,(IMG_{LGS})\,,\ Kurtosis\,(IMG$
    **end for** ▷ Image graph-based shortest path features
    $\{IMG_{Block}\} \leftarrow Partition(IMG)$
    **for** <$IB^i \in \{IMG_{Block}\}$> **do**
      $i \leftarrow 1\,to\,n; \{Vertex_G\} \leftarrow Graph_Mapping(IB^i)$
      **for** <$each\,(v_i, v_j\,) \in \{Vertex_G\}$> **do**
        $e_{ij} \leftarrow (v_i, v_j) \in A_M\,|\max\,(|x_i - x_j|\,, |y_i - y_j|) == T_e|$
        $\{Edge_G\} \leftarrow e_{ij}$
      **end for**
      **for** <$each\,e_{ij}\ \in\ \{Edge_G\}$> **do**
        $e_w \leftarrow abs\,\left(I_{v_i}(x_i, y_i) - I_{v_j}(x_j, y_j)\right) + \frac{I_{v_i}(x_i,y_i)+I_{v_j}(x_j,y_j)}{2}$
        $\{E_w\} \leftarrow e_w$
      **end for**
      $Identify S_v, D_v \in I_B foreach\,Path_A, A \in \{0^0,\ 45^0,\ 90^0, 135^0\}$
      $\{P_A\} \leftarrow Dijkstrasprocedure\,(S_v, D_v, I_B)$
      $\{P_{A_mean}\} \leftarrow mean\,\{P_A\}\ \,, IB^i_{A_mean} \leftarrow \{P_{A_mean}\}\,,\ i \leftarrow 1\,to\,n$
    **end for**
    **for** <$i\ from\,1\ to\ n$> **do**
      $\{Mean_A\} \leftarrow mean\{IB^i_{A_mean}\}$
    **end for**
    $GSP_{IMG} \leftarrow \{Mean_A\}$
    $Feature_vector_{IMG} = \{FoS_{IMG},\ SoS_{IMG},\ HoS_{IMG},\ HoC_{IMG},\ LGS_{IMG},\ GSP_{IMG}\}$
  **end for** 3
**end procedure**

## 4. Glaucoma image classification

WNN and MLP are utilized for glaucoma classification in this approach.

### 4.1. Wavelet neural network (WNN)

The structure and working principles of biological neurons are inherited by neural networks [50]. The most recent version of NN is WNN. The WNN incorporates both discrete wavelets and MLP learning capabilities to become a powerful classifier. The structural appearance of WNN is similar to an MLP structure that contains one input layer, one or more hidden layers, and one final output layer. In WNN, instead of non-linear activation functions (such as sigmoid), discrete wavelet functions are used for neuron activation as shown in Fig. 6.To construct the activation function of a neuron, the mother wavelet (*Mexican hat, Gaussian, Shannon, Morlet, Haar, GGW, and Meyer*) [51, 52] needs to be selected first. In the activation function, the wavelet coefficients recognize the supplied input significance and activate to get the predefined output in the training process.

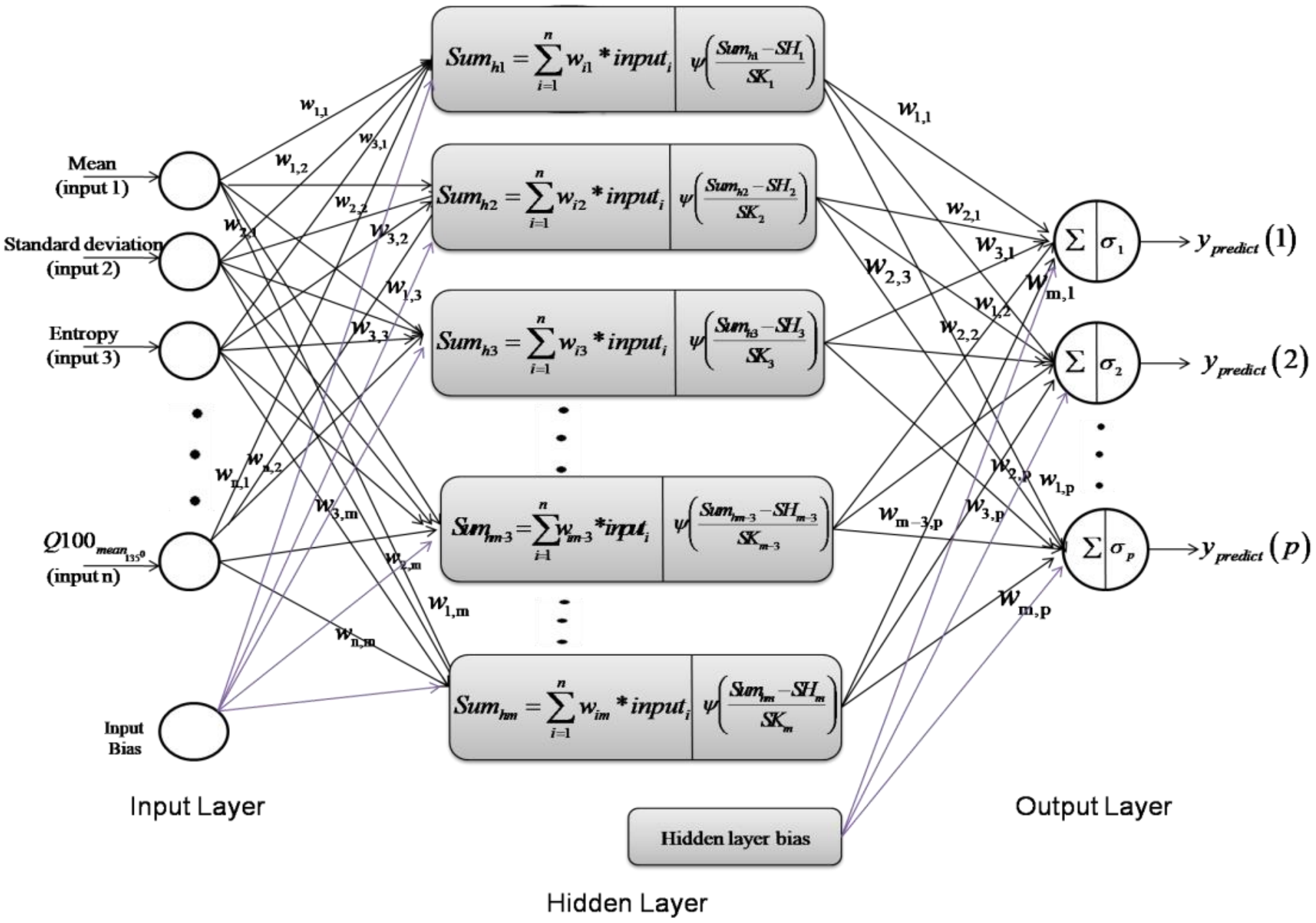

**Fig. 6 A three-layer wavelet neural network**

### 4.2. Multilayer perceptron neural network (MLP)

The learning is based on a back propagation function applied to a feed-forward network in the MLP neural network. The network is constructed with interconnections of artificial neurons to map input non-linearly with output. The general structure of MLP includes one input, one output, and one or more hidden layers [53]. In our experiment, the sigmoid activation function is used for the identification of glaucoma retinal images.

## 5. Result analysis

Our approach started with the phase of retinal image preprocessing, followed by the retinal image enhancement phase. Novel features are extracted from qualitative retinal images, which are fed to MLP and WNN classifiers to obtain satisfactory results that are discussed in this section.

Image preprocessing concentrates on improving the image brightness and contrast. Image brightness is improved using quantile-based histogram modification based on QRTM CDF transformation. This approach takes care of the over-enhancement of image brightness. We have compared the results of our preprocessed images with histogram equalization (HE) results. The HE over-enhances the image brightness, which leads to unwanted retinal image structural alterations. Improvement of brightness and contrast of the retinal image using our approach and the HE approach are qualitatively compared in Fig. 7a. At the same time, our method improves the brightness of the image while preserving the contrast difference between retinal structures. In the case of the HE method, both the optic disc and cup areas are filled with the same contrast and it becomes difficult to identify the borders of the cup and disc areas as visualized in Fig. 7b. The next phase of our approach is the enhancement of preprocessed retinal images to improve the edges of retinal structures.

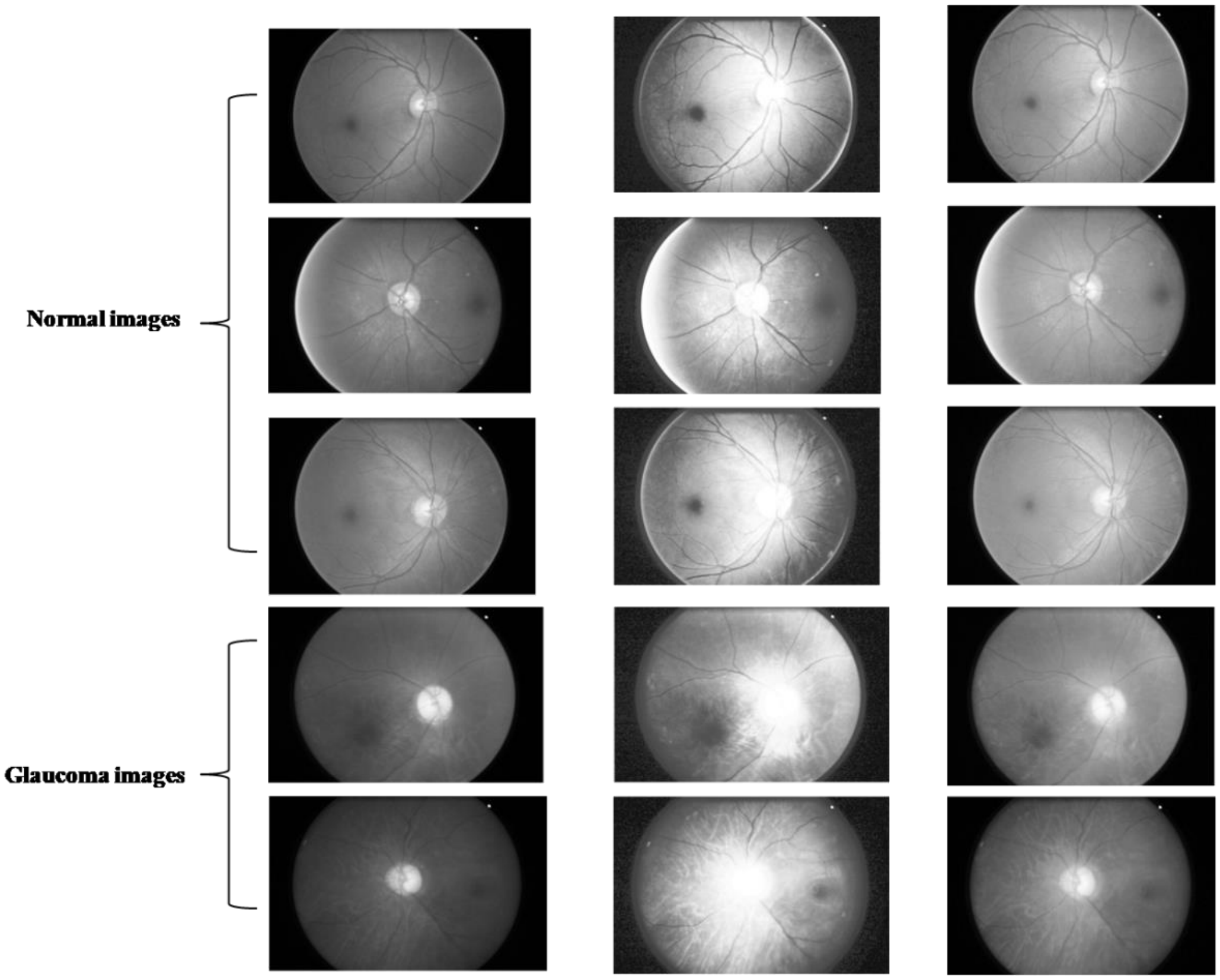

**(a) Original gray image** **(b) Images resulted by HE** **(c) Images resulted by the our preprocessing method**

**(a) Comparison of the HE and our proposed pre-processed results**

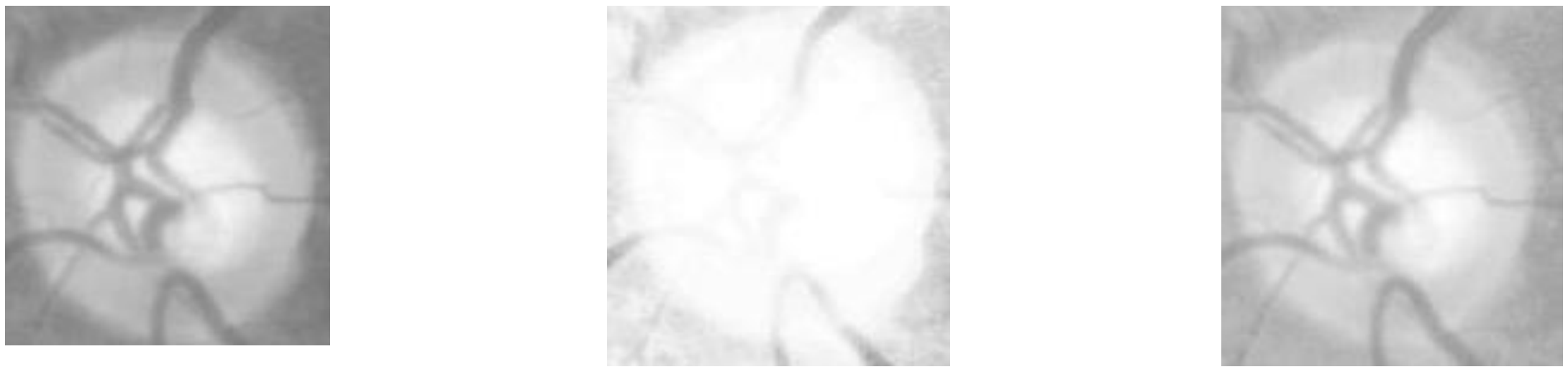

**(i) (ii) (iii)**

**(b) Optic disc and cup areas of (i) Original image, (ii) HE resulted image and (iii) proposed pre-processing resulted image.**

**Fig. 7: Proposed preprocessing method's results**

Our enhancement approach is based on independent modifications of low and high sub-band image DTCWT coefficients using $D_{SE}$. The value of *t* is dynamically changed with each image based on the s-curve transformation. In our approach, some images' enhancement stops at smaller values of *t*, and some images require larger *t* values, as shown in Fig. 8.

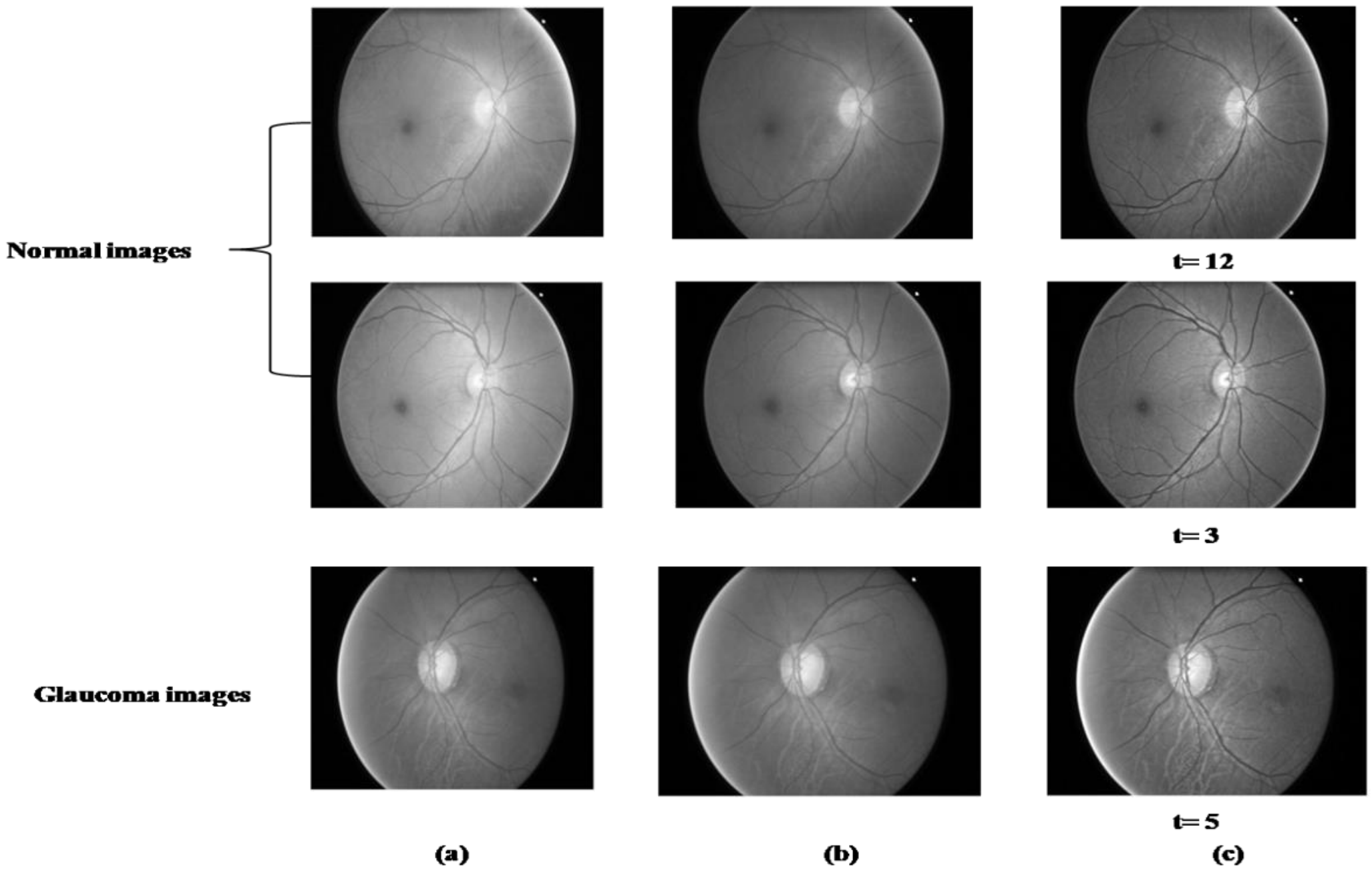

**Fig. 8: Illustration of dynamic structuring element usage: a) pre-processed image; b) enhanced image with t=0; and c) enhanced image with different t values of $D_{SE}$ in the proposed method.**

Increasing the *t* value above the value defined by the s-curve transformation does not show any improvement in the retinal structures. We have employed the mean square error (*MSE*) as a quality evaluation metric, and its lower value indicates the originality between the improved and the original images. MSE is calculated on our images resulted from the proposed method and compared the results with the HE, CLAHE, and RRIE [54] approaches as shown in Table 1.

**Table 1: Comparison of MSE obtained in our approach with other proposals**

| Sl. no | Image name | MSE | | | |
|---|---|---|---|---|---|
| | | HE | CLAHE | RRIE | Proposed method |
| 1 | Im0010_ORIGA | 0.067679 | 0.011415 | 0.002063 | 0.000715 |
| 2 | Im0012_ORIGA | 0.053374 | 0.008443 | 0.002022 | 0.00078 |
| 3 | Im0023_ORIGA | 0.044925 | 0.011075 | 0.000736 | 0.001336 |
| 4 | Im0026_ORIGA | 0.05219 | 0.008132 | 0.001127 | 0.000364 |
| 5 | Im0028_ORIGA | 0.038277 | 0.009291 | 0.001481 | 0.000542 |
| 6 | Im0145_ORIGA | 0.07223 | 0.013453 | 0.001212 | 0.000428 |
| 7 | Im0146_ORIGA | 0.034566 | 0.011431 | 0.001019 | 0.000436 |
| 8 | Im0206_ORIGA | 0.100455 | 0.021695 | 0.000792 | 0.00033 |
| 9 | Im0208_ORIGA | 0.066481 | 0.011787 | 0.001098 | 0.000588 |
| 10 | Im0618_g_ORIGA | 0.122118 | 0.019556 | 0.000544 | 0.000224 |
| 11 | Im0600_g_ORIGA | 0.073892 | 0.013195 | 0.002001 | 0.000708 |
| 12 | Im0586_g_ORIGA | 0.037967 | 0.013199 | 0.000924 | 0.000431 |
| 13 | Im0579_g_ORIGA | 0.058624 | 0.009337 | 0.001147 | 0.000295 |
| 14 | Im0510_g_ORIGA | 0.05132 | 0.011672 | 0.001403 | 0.000784 |
| 15 | Im0494_g_ORIGA | 0.074081 | 0.012381 | 0.000804 | 0.000273 |

The comparison results show that enhanced images resulting from our proposed preprocessing and enhancement approach have lower MSE values than other methods in the literature. The MSE histogram plot for RRIE and the proposed enhancement in Fig. 9 show that our approach keeps the images closer to the ground truth. For better visibility in the plot, HE and CLAHE results are not included in the plot. The enhanced retinal images are utilized to extract statistical and graph-based retinal image features to form a feature vector (Eq. 7). The input feature vector set is split into 75:25 ratios for training and testing /validation.

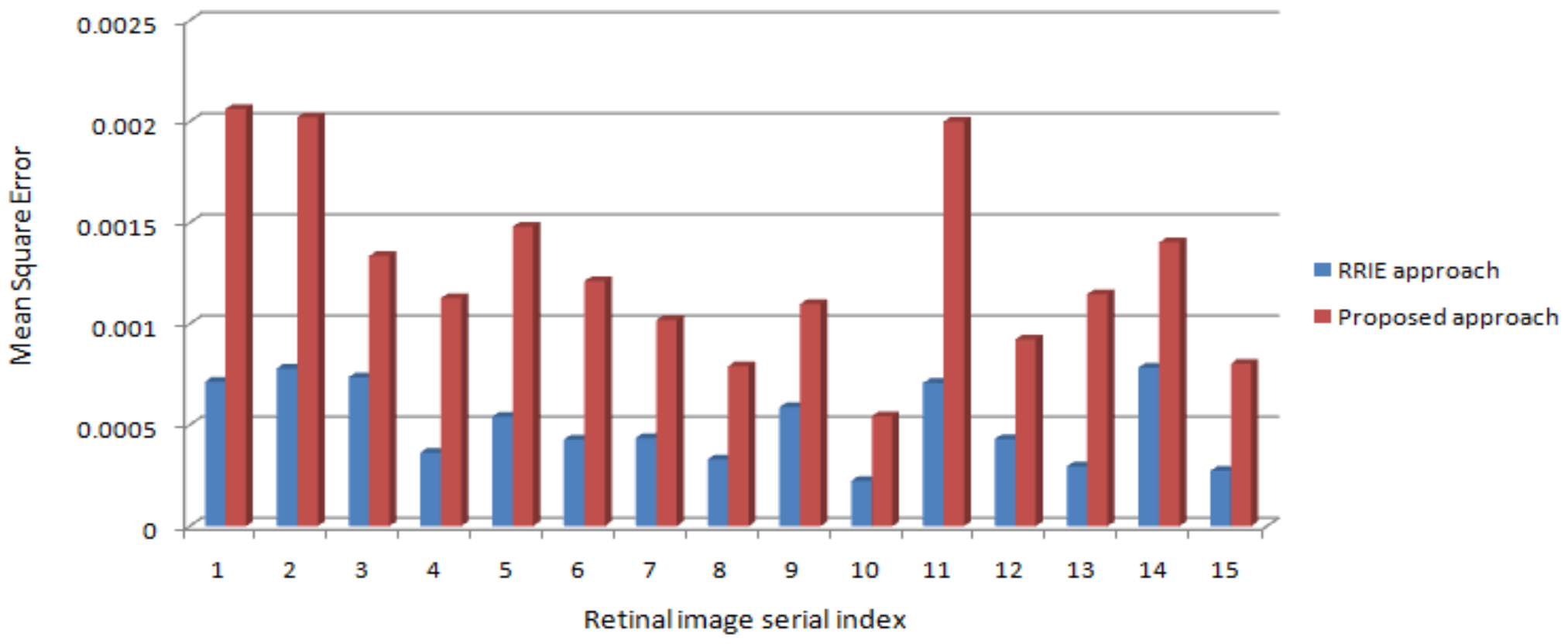

**Fig. 9 MSE histogram plot for result comparison**

$$Feature_vector = \left\{ \begin{array}{c} Mean, Standard_deviation, Entropy, Variance, Smoothness, \\ Kurtosis, Skewness, IDM, Contrast, Energy, Homogeneity, \\ HOC_{10^0}, HOC_{50^0}, HOC_{90^0}, HOC_{130^0}, HOC_{180^0}, \\ Entropy_{HoS}, Mean_{HoS}, Ent_{dg1}, Ent_{dg2}, Ent_{dg3} \\ Mean_{LGS}, Variance_{LGS}, Skewness_{LGS}, Kurtosis_{LGS}, Energy_{LGS}, Entropy_{LGS} \\ Kurtosis_{i^0}, Skewness_{i^0}, SD_{Mean_{i^0}}, Q25_{Mean_{i^0}}, Q50_{Mean_{i^0}}, Q75_{Mean_{i^0}}, Q100_{Mean_{i^0}} \ldots i \in \{0,45,90,135\} \end{array} \right\} \quad (7)$$

We have carried out classification by changing the number of nodes in the hidden layer to find the optimum number of nodes in the hidden layer. We also conducted tests to fix the network regulation parameters and number of epochs to get optimum performance. In our approach, the WNN performance is measured in terms of the testing error rate. In the training processes of WNN and MLP, the network parameters are updated using batch-wise validation and training sets. As shown in Table 2, we selected the wavelet activation function based on the testing error (TE) rate of the WNN for 50 epochs.

**Table 2: Testing error for various mother wavelet functions**

| S. No | Wavelet function | Testing error |
|---|---|---|
| 1 | Mexican | **3.6721 %** |
| 2 | Mayer | 9.4129% |
| 3 | Gaussian | 9.1360% |
| 4 | GGW | 10.2434% |
| 5 | Morlet | 9.9666% |

WNN with Mexican hat wavelet activation function performs best with given retinal image features as it achieves lower testing. In our approach, WNN performance is measured in terms of testing error percentage, and we have obtained it as 3.65% for 200 epochs. We have also implemented the MLP network for glaucoma classification and obtained a 9.19% test error for 200 epochs. Performance graphs are shown in Fig. 10.

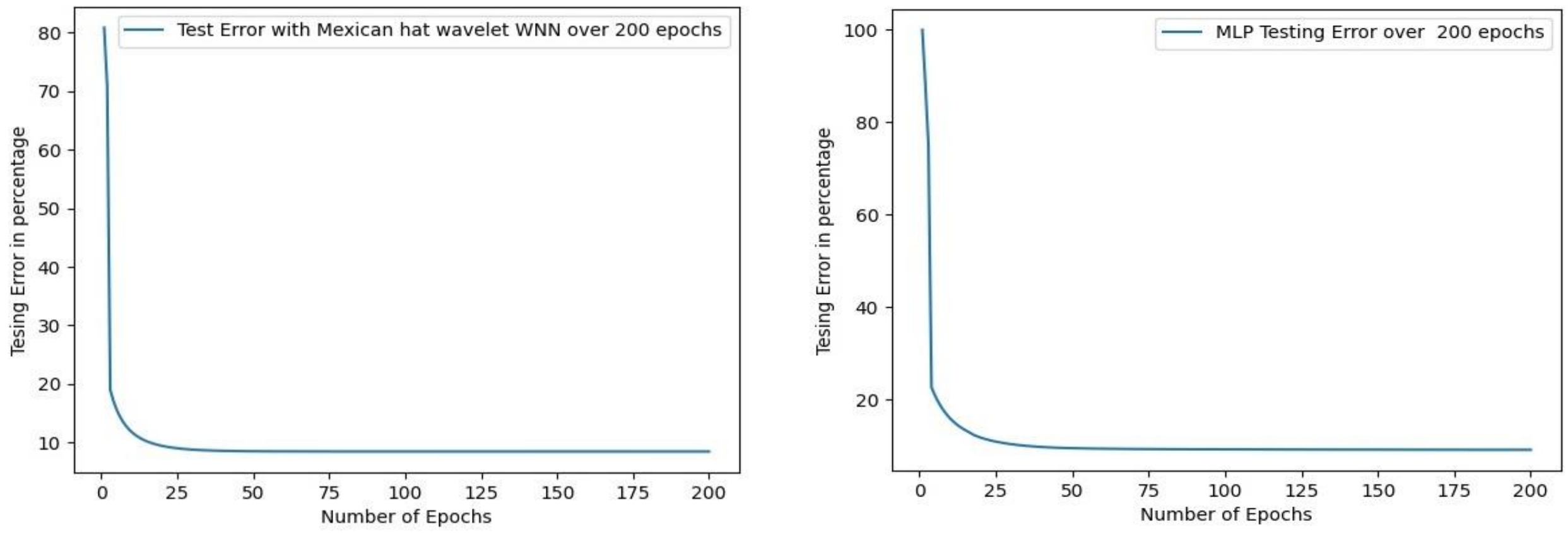

**(a) WNN classification** **(b) MLP classification**

**Fig. 10 WNN vs. MLP glaucoma classification performance plots**

In our experiment, it has been observed that WNN performance is superior to MLP for the glaucoma classification. The sigmoid activation function is used in MLP with 24 hidden neurons. However, WNN achieved a lower testing error with 10 hidden neurons using the Mexican hat wavelet activation function. To carry out an in-depth WNN and MLP performance comparison, the retinal image classification is done with the extracted feature sets by varying the number of hidden nodes, epochs, and regularization parameters. The obtained results are shown in Table 3.

**Table 3: WNN performance comparison with MLP**

| HU | BS | Epochs | NN | $V_{error}$ (%) | $T_{error}$ (%) | NN | $V_{error}$ (%) | $T_{error}$ (%) |
|---|---|---|---|---|---|---|---|---|
| 05 | 113 | 10 | Wavelet neural network | 7.757744 | 9.576151 | Multilayer perceptron neural network | 11.736765 | 16.116977 |
| | | 50 | | 7.723564 | 8.582367 | | 11.921193 | 13.209610 |
| | | 100 | | 7.723564 | 8.582367 | | 11.845652 | 13.179410 |
| | | 150 | | 7.723563 | 8.582126 | | 11.784855 | 13.199261 |
| | | 200 | | 7.723563 | 8.582126 | | 11.705995 | 13.159797 |
| 10 | 56 | 10 | | **3.3829828** | **5.743336** | | 13.640801 | 22.358791 |
| | | 50 | | **3.2350709** | **3.672188** | | 11.633276 | 13.302588 |
| | | 100 | | **3.2350360** | **3.655132** | | 11.442654 | 12.942912 |
| | | 150 | | **3.2350359** | **3.655034** | | 11.367036 | 12.847224 |
| | | 200 | | **3.2350359** | **3.655033** | | 11.345857 | 12.815169 |
| 15 | 37 | 10 | | 8.275527 | 14.14786 | | 15.346539 | 28.207156 |
| | | 50 | | 7.456985 | 8.806679 | | 10.994153 | 14.284100 |
| | | 100 | | 7.454990 | 8.586176 | | 10.840125 | 12.652520 |
| | | 150 | | 7.454983 | 8.579782 | | 10.817564 | 12.601801 |
| | | 200 | | 7.454982 | 8.579562 | | 10.796918 | 12.663610 |
| 24 | 23 | 10 | | 9.121266 | 17.03973 | | **9.785792** | **16.040003** |
| | | 50 | | 7.475102 | 9.520537 | | **8.345611** | **9.543162** |
| | | 100 | | 7.448297 | 8.737305 | | **8.208860** | **9.285133** |
| | | 150 | | 7.447930 | 8.654753 | | **8.154612** | **9.216487** |
| | | 200 | | 7.447930 | 8.654753 | | **8.139419** | **9.193491** |

HU: Hidden units; BS: Batch size; $V_{error}$: validation error; $T_{error}$: Testing error

The WNN achieves minimal testing error almost at the 50$^{th}$ epoch, irrespective of the number of hidden nodes. However, MLP achieves the lowest testing error with an increasing number of hidden nodes as well as the number of epochs. The proposed approach is also applied on the other public retinal datasets (ACRIMA and DRISHTI) and results are shown in Table 4.

Both the WNN and MLP classifiers are applied to the extracted graph-based and statistical features before and after image enhancement. The classifiers' performance has considerably improved with the application of the proposed image preprocessing and enhancement approach before feature extraction, which is shown in Fig. 11. In the case of ORGIA and ACRIMA datasets, the WNN has achieved 96.36% and 98.24% accuracy, which are far better than the accuracies achieved by MLP, i.e., 90.90% and 91.22%. However, both WNN and MLP have achieved less accuracy, i.e., 85.71% with the DRISHTI due to its smaller dataset size. Before retinal image enhancement, WNN classifies the images of ORGIA and ACRIMA with accuracies of 92.72% and 94.73%, whereas MLP classifies them with accuracies of 89.09% and 91.22%. The performance of WNN is constant with the DRISHTI dataset before and after enhancement, but the MLP performance is degraded in the absence of image enhancement.

| ML Classifier | Approach | DS | No. of images | Accuracy (%) | Sensitivity (%) | Specificity (%) | Confusion matrix (%) | | | |
|---|---|---|---|---|---|---|---|---|---|---|
| | | | | | | | $T_P$ | $F_P$ | $T_N$ | $F_N$ |
| **WNN** | Before image enhancement | O | 650 | 92.72 | 100 | 83.33 | 1.0 | 0.17 | 0.83 | 0 |
| | | A | 705 | 94.73 | 100 | 88.88 | 1.0 | 0.12 | 0.88 | 0 |
| | | D | 101 | 85.71 | 100 | 83.33 | 1.0 | 0.17 | 0.83 | 0 |
| **MLP** | | O | 650 | 89.09 | 100 | 75.00 | 1.0 | 0.25 | 0.75 | 0 |
| | | A | 705 | 91.22 | 100 | 81.48 | 1.0 | 0.19 | 0.81 | 0 |
| | | D | 101 | 71.42 | 100 | 66.66 | 1.0 | 0.34 | 0.66 | 0 |
| **WNN** | After image enhancement | O | 650 | 96.36 | 100 | 91.66 | 1.0 | 0.09 | 0.91 | 0 |
| | | A | 705 | 98.24 | 100 | 96.29 | 1.0 | 0.04 | 0.96 | 0 |
| | | D | 101 | 85.71 | 100 | 83.33 | 1.0 | 0.17 | 0.83 | 0 |
| **MLP** | | O | 650 | 90.90 | 100 | 79.16 | 1.0 | 0.21 | 0.79 | 0 |
| | | A | 705 | 91.22 | 100 | 81.48 | 1.0 | 0.19 | 0.81 | 0 |
| | | D | 101 | 85.71 | 100 | 83.33 | 1.0 | 0.17 | 0.83 | 0 |

*DS: Dataset; O: ORGIA; A: ACRIMA; D: DRISHTI

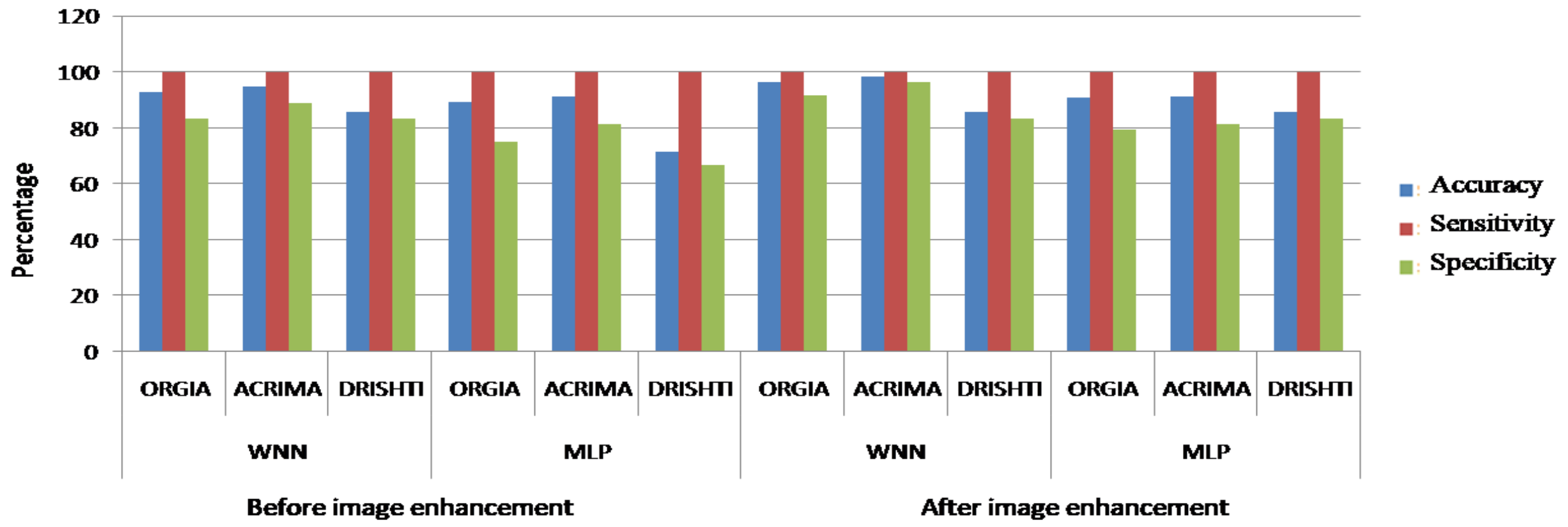

**Fig. 11 WNN and MLP performance plots for various datasets**

Both the WNN and MLP classifiers are applied to the extracted graph-based and statistical features before and after image enhancement. The classifiers' performance has considerably improved with the application of the proposed image preprocessing and enhancement approach before feature extraction, which is shown in Fig. 13. In the case of ORGIA and ACRIMA datasets, the WNN has achieved 96.36% and 98.24% accuracy, which are far better than the accuracies achieved by MLP, i.e., 90.90% and 91.22%. However, both WNN and MLP have achieved less accuracy, i.e., 85.71% with the DRISHTI due to its smaller dataset size. Before retinal image enhancement, WNN classifies the images of ORGIA and ACRIMA with accuracies of 92.72% and 94.73%, whereas MLP classifies them with accuracies of 89.09% and 91.22%. The performance of WNN is constant with the DRISHTI dataset before and after enhancement, but the MLP performance is degraded in the absence of image enhancement.

By considering several aspects of the obtained results, the WNN performance for glaucoma classification with the proposed images' extracted features is much better than MLP networks due to the combination of time-frequency localization of wavelets and self-learning capability.

## 6. Conclusion

Our proposed technique contributes to all major CAD phases, including retinal image preprocessing, enhancement, feature extraction, and classification. Our approach initially concentrated on retinal image quality enhancement that is carried out in two phases: preprocessing and enhancement. The preprocessing improves the retinal image brightness and contrast using QRTM-transformed CDF-based image quantile histogram modification. Our preprocessing method ensures that the brightness and contrast are optimized. Image enhancement improves the retinal structures using an independent modification of the retinal images' DTCWT low and high-pass sub-band coefficients. Retinal image texture patterns are captured using graph-based retinal image features using proposed local graph structures and graph-based shortest paths in various directions, along with statistical features. The glaucoma classification is carried out using WNN and MLP networks by selecting activation functions systematically. In our approach, WNN performs better than MLP for glaucoma classification. The classification results are tabulated for various network parameter combinations. According to the available literature, our strategy is a fresh approach that combines retinal images' graph-based feature extraction with the WNN-based glaucoma classification. In the future, our research will continue to identify a new version of graph-based retinal image features for glaucoma classification with multi-level WNN.